\documentclass[twocolumn]{article}
\usepackage{graphicx}
\usepackage{amsmath}
\title{Algorithmic Detection of Computer Generated Text}
\author{Allen Lavoie -- lavoia@rpi.edu\\
        Mukkai Krishnamoorthy -- moorthy@cs.rpi.edu\\
        Rensselaer Center for Open Source Software (RCOS)\\
        Rensselaer Polytechnic Institute}
\begin{document}
\date{}
\maketitle
\section{Abstract}
Computer generated academic papers have been used to expose a lack of thorough human review at several computer science conferences.  We assess the problem of classifying such documents.  After identifying and evaluating several quantifiable features of academic papers, we apply methods from machine learning to build a binary classifier.  In tests with two hundred papers, the resulting classifier correctly labeled papers either as human written or as computer generated with no false classifications of computer generated papers as human and a 2\% false classification rate for human papers as computer generated.  We believe generalizations of these features are applicable to similar classification problems.  While most current text-based spam detection techniques focus on the keyword-based classification of email messages, a new generation of unsolicited computer-generated advertisements masquerade as legitimate postings in online groups, message boards and social news sites.  Our results show that taking the formatting and contextual clues offered by these environments into account may be of central importance when selecting features with which to identify such unwanted postings.

\section{Introduction}
A project called Scigen\cite{scigen} made waves in 2005 when it produced a paper which was accepted to WMSCI 2005.  Unfortunately this was not due to breakthroughs in artificial intelligence which allowed Scigen to write about computer science effectively.  Instead, Scigen relies on sentence generation based on a context free grammar.  It effectively produces an assortment of random keywords spliced into predefined sentence structures.  With the addition of randomly generated graphics and references, the papers are practically indistinguishable from papers that humans have written until true understanding is attempted by someone who could reasonably expect to understand an equivalent but meaningful paper.  The question then arises as to whether a human review is required to classify such papers.

The difficulty of this problem is quite dependent on the approach used and that approach's expected applicability.  On one extreme we could look for formatting quirks or other metadata which are irrelevant to the paper itself, but which could easily be used to identify Scigen papers specifically.  This approach has very little general applicability, but would be almost trivial to implement.  Another extreme is the most general case of identifying papers as either examples of good scholarly work or not.  A solution here would be of great practical use, but seems unlikely at the present time.  A good approach, then, is one which maximizes its general applicability while remaining practical.  

This paper attempts to show that there exists a non-empty subset of approaches ranging from trivial to hopelessly complex which are both practical and useful in a broader context.  While practicality has objective measures such as algorithmic complexity, the usefulness of an approach is necessarily subjective.  Thus we will analyze one approach for practicality and effectiveness in dealing with Scigen papers specifically, and then discuss other potential applications.

\subsection{General Approach}
Having decided not to exploit shallow flaws in Scigen and lacking any true understanding of the academic papers we seek to classify, we draw candidate features from several simple observations about human writing.
\begin{enumerate}
\item
If a paper claims to be about a specific topic, it should actually be about that topic.
\item
Papers follow a central theme.
\item
Cited papers are related to the paper which cites them.
\end{enumerate}
Unfortunately none of these observations are directly quantifiable as stated.  Note that even the above criteria are necessary but not sufficient for identifying good scholarly work.  They do, however, suggest heuristics which can easily be quantified.  We can rewrite each of these observations in terms of keywords, sacrificing accuracy for computability.  
\begin{enumerate}
\item
  Keywords which appear in the title and abstract of a paper should appear frequently in the body of that paper.
\item
  Certain keywords should be favored throughout a paper.
\item
  A paper should mention keywords from the titles of articles it cites.
\end{enumerate}
Whereas our original observations capture part of what it means for a paper to be an example of true scholarly work, the keyword heuristics are merely plausible substitutes.  On the other hand, these features are readily quantifiable.  Since our features are no longer directly tied to the definition of scholarly work, a paper generator could easily be adapted to superficially fulfill the above heuristics.  The simplicity of keyword heuristics in general, however, also allows us to readily produce additional features.  For example, we could examine repetition within paragraphs or patterns in the usage of certain sentence structures.  Domain-specific features, such as the occurrence of keywords from a user-submitted summary in a linked article on a social news website, are also feasible.

Having a basis for quantifying the abstract concept of scholarly work, we can now build a classifier and test our intuitive notions empirically.  
\section{Algorithm}
\subsection{Preprocessing}
We begin by converting a subject paper to plain text from the typical PDF.  There are several freely available tools for accomplishing this task.  Next, a paper is split into word tokens.  

Before additional processing is done to clean up the text, we search for certain keywords denoting the title, abstract, introduction and references sections of a paper.  Keeping such structure allows us to base features on comparisons between sections, and doing it early ensures that our text processing does not interfere.  In the case of missing keywords, we simply fall back to taking a certain fraction of the paper at the desired section's expected location.  

Next, we clean the text by selecting only parts of speech which might reasonably have keywords related to the topic of the paper.  This entails standard part of speech tagging followed by aggressive filtering.  Specifically we selected nouns, adjectives and foreign or unrecognized words.  Words which were unrecognized by the tagging algorithm often turned out to be the most valuable, as many times they were examples of unique technical jargon.  

The selected tokens are finally stemmed to avoid confusion between different forms of a single word.  This allows for a straightforward character based comparison, which turned out to be good enough for our purposes.  A more advanced approach might make use of a semantic difference metric \cite{semantic_diff}. 
\subsection{Feature scoring}
The text can then be scored numerically based on our selected features.  We need to make the three features discussed in the previous section slightly more specific in order to accomplish this.  Again we trade some semantics for ease of computation as we introduce fairly arbitrary nuances to our features, but this time the sacrifices are fairly subtle.  

\subsubsection{Title and abstract score}
For our first feature, we could perform a straightforward count of keywords from the title and abstract of a paper in that paper's body.  This is a decent first approximation, but favors long papers unduly.  Thus, we normalize this feature by scoring papers based on the number of times keywords from the title or abstract of a paper are mentioned divided by the length of the part of speech filtered body of that paper.  Let $A$ be the set of keywords from the title and abstract of a paper after part of speech filtering, and $B$ be the corresponding multiset for the remainder of the paper similarly filtered.  Here, $m_Q(q)$ denotes the multiplicity of element $q$ in the multiset $Q$.  Then our first feature's score $s_1$ can be written as follows:
\begin{align}
  s_1 = \frac{\sum_{a \in A} m_B(a)}{|B|}
\end{align}
We must be careful to treat $B$ as a multiset and not simply as a set, since this feature seeks to quantify the repetition of ideas from the title and abstract of a paper in its body.  We could similarly take the repetition of keywords from the title and abstract of a paper into account by making $A$ a multiset as well, but doing this simply rewards repetition rather than completeness.  Whereas the abstract and title of a paper taken together should be a succinct summary of the paper's content, rephrasing and repetition is both common and desirable in a paper's body.  

\subsubsection{Word repetition score}
Next, we seek to quantify the repetition of a certain set of words throughout a paper.  Here we have chosen to compare the occurrence of the top $N$ most used words in a paper to the occurrence of all other words.  Let $P$ denote the multiset of all words in a given paper, and let $W=\{w_i\}$ denote the set of distinct elements in $P$ sorted in decreasing order of their multiplicity $m_P(w_i)$.  Thus $w_0$ occurs with the highest frequency in the paper, followed by $w_1$ and so on.  Here $|P|=\sum_{w_i \in W}m_P(w_i)$.  Then the second feature can be written as follows:
\begin{align}
  s_2 = \frac{\sum_{i = 0}^{N-1}m_P(w_i)}{|P| - \sum_{i = 0}^{N-1}m_P(w_i)}
\end{align}
In general $N$ must be between $1$ and $|W|-1$ inclusive, but we have found that $N=10$ is a fairly good trade off for this feature.  Note that part of speech tagging and filtering plays a very important role here.  Without it we would almost certainly be selecting words which are simply common in the English language, at which point our feature would cease to make sense.

\subsubsection{References score}
Our final feature is calculated much like the first.  Without parsing references at all, we simply use the tokenized, filtered and stemmed set of keywords from the references section.  This includes paper titles, authors and a lot of other irrelevant information.  The irrelevant tokens do not affect the feature's score since we do not normalize on the number of tokens in the references section.

If we let $R$ denote the set of tokens from the references section of a paper and again let $B$ denote a multiset of keywords in the remainder of the paper, we can compute the third feature as follows:
\begin{align}
  s_3 = \frac{\sum_{r \in R} m_B(r)}{|B|}
\end{align}
While feature three's computation is nearly identical to (1), its semantics are quite different.  While our first feature attempts to quantify the relevance of the title and abstract of a paper, our third feature seeks to quantify the relevance of its chosen references.  Thus despite the similar computation, we do not see a strong linear dependence between these two features.

\subsection{Classification}

Finally, we build a classifier based on all three features.  Let a paper $p$ be represented by a point $(s_1, s_2, s_3)$ in a three dimensional space, where $s_1$, $s_2$ and $s_3$ are our first, second and third features respectively.  We can then build a classifier based on one of several well known methods from machine learning.  A nearest neighbor classifier\cite{knn} which takes a vote of the $k$ nearest points ($k=3$ in this case) with known classifications was used for its simplicity on small data sets for the purposes of this paper, but support vector machines\cite{svm} or other more advanced algorithms would be more efficient when dealing with larger data sets.

\subsection{Running time}

The running time of our implementation of the algorithm outlined above is dominated by the part of speech tagging used during preprocessing.  We use the default part of speech tagger from the Natural Language Tool Kit\cite{nltk}, but a faster and slightly less accurate tagging algorithm could dramatically reduce the running time of the algorithm presented above.  Tagging accuracy should not affect the accuracy of the overall algorithm overmuch, since we are simply using the tags to accept or reject word tokens.  

After preprocessing, feature scores can be generally be calculated sub-quadratically in the length of a paper by using either search trees or hash tables for calculating the multiplicity of a given keyword.  In practice, this step is quite fast compared to preprocessing.  

Finally, classification relies on one of several well known binary classifiers.  For simplicity, we used a nearest neighbor search with a KD-tree\cite{knn_algs} based on a Euclidean distance metric.  A support vector machine or other classifier whose running time during classification is independent of the size of the set of training data can be substituted where efficiency is a concern.

\section{Results}
After scoring 200 papers based on the above features, a 3-nearest-neighbor classifier misclassified only 2 papers for an error rate of 1\%.  For error estimation, we used leave one out cross validation.  The data set consisted of 100 computer generated papers from Scigen and 100 randomly chosen papers from the computer science and mathematics sections of the ArXiv\cite{arxiv}.  

Of the two misclassified papers, the first\cite{second_error} has an exceptionally short abstract and a great deal of formulas.  The short abstract yields a low score for feature 1, since it mentions very few relevant keywords.  The formulas did not translate well to text, and our failure to filter out the artifacts of this translation made for a slightly reduced score for feature 2.  The second\cite{first_error} paper had no exceptionally low score, but simply did not score well on any feature, placing it well within a cluster of computer generated papers.  Neither paper appears to be computer generated upon human inspection.

\begin{figure}
\includegraphics[width=\linewidth]{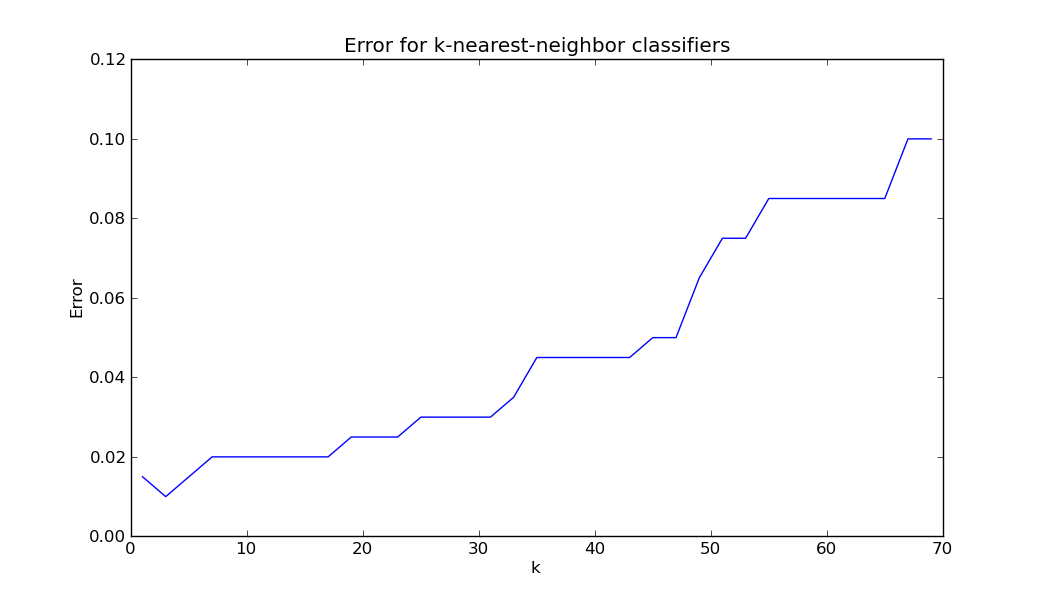}
\caption{Error rates for nearest neighbor classifiers of order $k$ using leave one out cross validation.}
\label{fig:error_rates}
\end{figure}

Notably, we did not have any false negative classifications.  While human papers showed a great deal of variation leading to the errors mentioned above, Scigen papers fell within fairly well defined ranges on each feature.  

We mention above that our error was measured with a 3-nearest-neighbor classifier.  This is perhaps somewhat surprising, since the model provides almost no regularization and we do none outside of the model.  However, Figure \ref{fig:error_rates} indicates that this level of regularization is preferable to that found in higher order nearest neighbor classifiers.  The same general trend is visible when pruning is employed, and so we do not believe that the trend is simply an artifact of density effects.  

\begin{figure}
  \includegraphics[width=\linewidth]{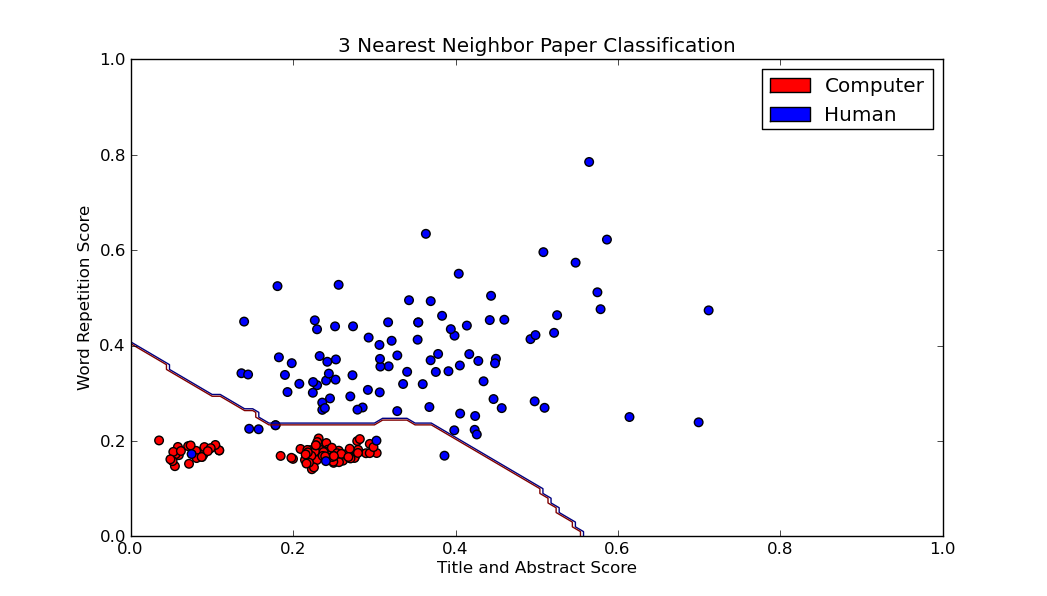}
  \caption{A two dimensional cross section of the classifier, ignoring the references score.}
  \label{fig:classifier}
\end{figure}

Figure \ref{fig:classifier} shows the distribution of papers and the classification boundary based on our 3-nearest-neighbor classifier and two of the three features: Word repetition and title and abstract scores.  The image is somewhat misleading, as the classifier works with all three features in a three dimensional space.  All but two of the points which appear misclassified in Figure \ref{fig:classifier} are differentiated by their references score.

The paper your are now reading was determined to be a human product by the 3-nearest-neighbor classifier.  

\section{Conclusions}
We have shown that the problem of computer generation of text is not quite so simple as stringing together keywords from a predefined distribution.  Coherent human writing has many subtly self-referential elements which can be exploited to classify products of unwary text generators.  Relying simply on keyword-based features, it is possible to exploit these elements efficiently to attack the problem of paper classification with only moderately sized data sets.

Traditionally, attempts to filter out automated messages have focused on their keyword composition rather than their structure.  Such techniques are ideal when very little context is available to the message generator, such as in the case of a new email message arriving.  Previously received messages provide a good deal of context for the classifier, but are unavailable when generating the unsolicited messages.  These techniques are less effective when a context is readily available to all parties, where the keyword composition of generated messages can be carefully selected to avoid detection.  

The structural elements described in this paper are certainly not difficult to duplicate in a text generator.  When generating papers, one must simply favor words chosen previously or in certain sections of the text.  However, the specific features discussed in this paper only scratch the surface of possible structural considerations when building classifiers.  As text generation evolves, text classifiers have many aspects of text structure from which to draw new heuristics.  Many of these heuristics will be domain specific, just as some of the features discussed in this paper are applicable only to academic papers.  

As user generated content becomes more prevalent, there is an increasing monetary incentive to pass unsolicited and automated messages off as human content in information sharing networks.  In many cases is it undesirable or infeasible to have all such messages moderated by a human, in which case techniques from machine learning such as those described in this paper will become increasingly expedient.

\subsection{Future Work}

It is worth considering the limits of text classifiers in general.  Consider a perfect binary classifier of papers as either scholarly work or not.  We could then construct a program to enumerate all examples of scholarly work by filtering an enumeration of the set of all strings.  This is certainly no proof of impossibility, but does seem unlikely.  It is then natural to ask how close a classifier can reasonably get to the aforementioned ideal, or even how effective a specific classifier is.

In this paper we perform a straightforward character based comparison on stemmed words.  There has been some work\cite{semantic_diff} in determining semantic differences between words.  We could then ask how a word relatedness heuristic in place of a simple character comparison affects a text classifier's accuracy.

Is automatic deep reference checking feasible, and if so can it make a reference score more effective?  There are automated tools which index academic papers and analyze citation networks, providing information which could be very useful in creating additional features for classifying computer generated academic papers specifically.

We propose quantifying the effectiveness of text classifiers as an open question above.  An analogous question can be asked about text generation.  Can we quantify the difficulty of classifying text generated by certain means?  Does the availability of a corpus of human texts make text generation with plausible structure easier?

\section{Source code}
The source code for an implementation of the algorithm discussed in this paper is available online at \texttt{http://code.google.com/p/paper-detection/}.

\section{Acknowledgments}
This work was made possible by the generous support from Sean O'Sullivan (RPI '85) of the Rensselaer Center for Open Source Software.  
\section{References}
\renewcommand\refname{}
\vspace*{-3em}
\bibliographystyle{plain}
\bibliography{algorithmic_detection}
\end{document}